\documentclass{article}
\usepackage{spconf,amsmath,graphicx}
\usepackage{hyperref}
\usepackage{algorithmic}
\usepackage{subfigure}
\usepackage[font=small,skip=3pt]{caption}
\usepackage[linesnumbered,ruled,vlined]{algorithm2e}
\SetKwInput{KwInput}{Input}                
\SetKwInput{KwOutput}{Output}              

\title{Attentive Dual Stream Siamese U-net for Flood Detection on Multi-temporal Sentinel-1 Data}
\name{Ritu Yadav,\thanks{The research is part of the project ‘EO-AI4GlobalChange’ funded by Digital Futures.}, Andrea Nascetti, Yifang Ban}
\address{Division of Geoinformatics, KTH Royal Institute of Technology, Sweden}
\begin{document}
%
\maketitle
\begin{abstract}
Due to climate and land-use change, natural disasters such as flooding have been increasing in recent years. Timely and reliable flood detection and mapping can help emergency response and disaster management. In this work, we propose a flood detection network using bi-temporal SAR acquisitions. The proposed segmentation network has an encoder-decoder architecture with two Siamese encoders for pre and post-flood images. The network's feature maps are fused and enhanced using attention blocks to achieve more accurate detection of the flooded areas. Our proposed network is evaluated on publicly available Sen1Flood11~\cite{bonafilia2020sen1floods11} benchmark dataset. The network outperformed the existing state-of-the-art (uni-temporal) flood detection method by 6\% IOU. The experiments highlight that the combination of bi-temporal SAR data with an effective network architecture achieves more accurate flood detection than uni-temporal methods.

\end{abstract}

\section{Introduction}
\label{sec:intro}

Natural disasters cost billions of dollars worth of economy every year, and floods are responsible for a major part of that. Because of floods, millions of people abandon their property, moreover poor and middle-class people are most affected by floods. The loss of lives and property due to natural disasters brings people towards poverty and it takes them decades to recover. With climate change, the developed countries are also at high risk. Prediction of floods and evacuation before the event is not quick enough and still improving. In such a scenario, accurate and reliable flood mapping after the disaster can help in rescue missions, re-routing traffic, delivering aids, and many more.

Satellites are a leading technology in gathering quick information on a large scale. Compared to optical data, Synthetic Aperture Radar (SAR) imagery is preferred for flood mapping from space. Unlike optical sensors, SAR has the capability of imaging day and night, irrespective of the weather conditions. 

SAR data acquired from various satellites has been explored for water detection and flood mapping. Historically, Flood mapping on SAR data is performed using manual thresholding, fuzzy logic, difference images, filtering, log-ratio, and others ~\cite{SARreview, Vassileva2015}. More recently, multiple studies investigated the potential of Deep Learning algorithms for the flood detection task, mainly using uni-temporal data. For example, in~\cite{asaro2021learning} work, authors experimented on support vector machine and basic neural networks. In paper ~\cite{amitrano2018unsupervised} and ~\cite{jeon2021water}, different flood events are considered and water segmentation is conducted on uni-temporal data using U-Net architecture.
Moreover, previous works investigated DL networks on smaller sites, hence lacking training data and generalization. In 2020, a large-scale flood dataset Sen1Flood11~\cite{bonafilia2020sen1floods11} was launched as a free and open benchmark dataset helping researcher in experimenting DL methods in flood detection tasks. This dataset has been explored in some studies ~\cite{akiva2021h2o, konapala2021exploring} and ~\cite{bai2021enhancement}. In study~\cite{akiva2021h2o}, authors experimented on optical Sentinel-2 data for the domain adaptation and flood segmentation task. In ~\cite{konapala2021exploring} and~\cite{bai2021enhancement}, SAR and optical data are fused to segment flood areas on uni-temporal data.

\begin{figure*}  
\centering  
\begin{subfigure}
  \centering  
  \includegraphics[trim=1cm 0.2cm 1cm 1cm, width=130mm, height=85mm]{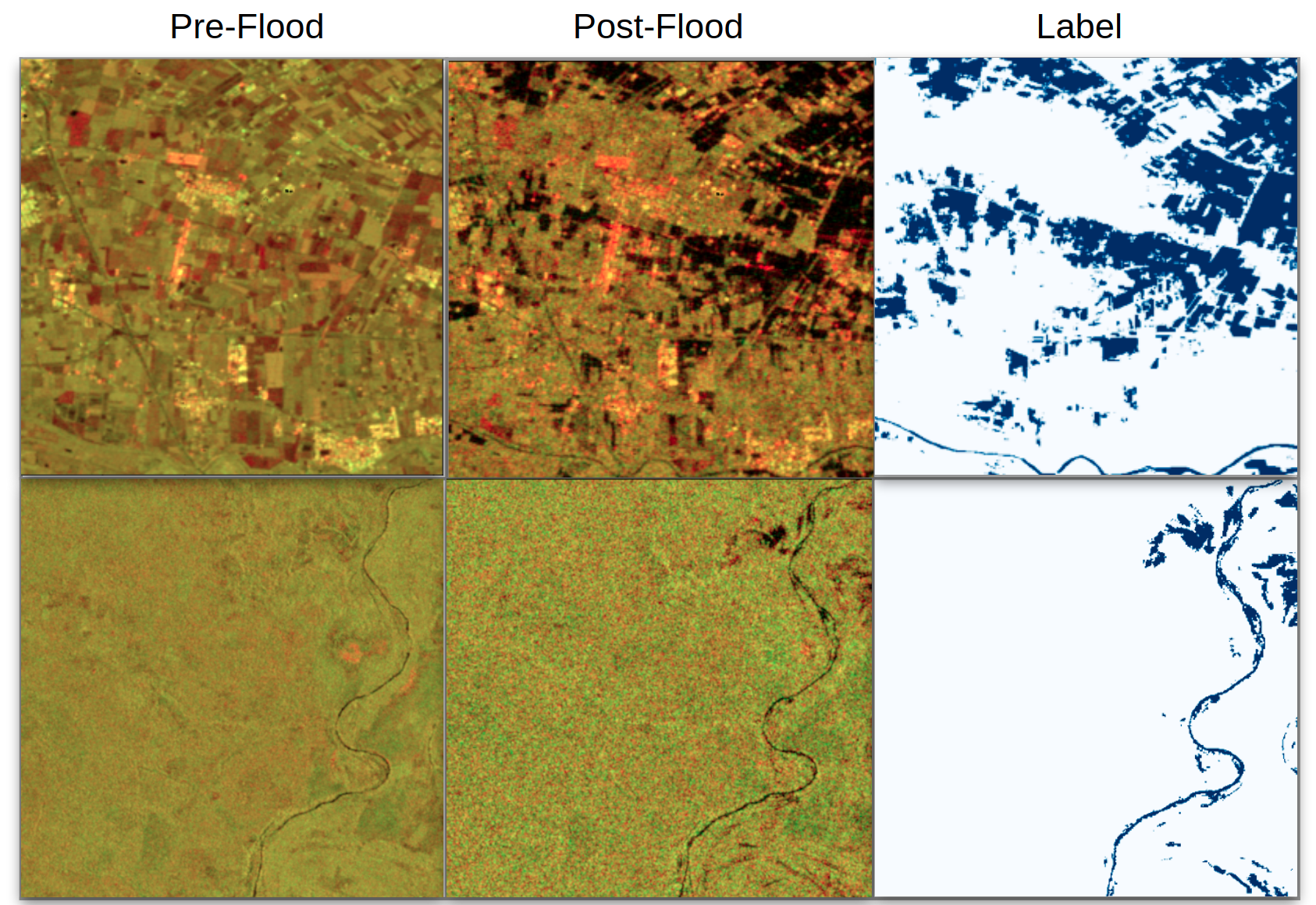}
\end{subfigure}
\caption{Data Samples. From left to right, pre-flood, post-flood images and Ground truth labels are visualized. In the ground truth blue color indicates water and the background is in white.}
\label{fig1}
\end{figure*}

In this work, flood detection is performed as a change detection task on bi-temporal data. For the experiments, we used the post-flood SAR images from Sen1Flood11~\cite{bonafilia2020sen1floods11} and pre-flood images are collected separately. In Fig. \ref{fig1}, an example of the input pre and post images and the reference label map are illustrated. 

\section{Proposed Method}
\label{sec:method}

\subsection{Data Preparation}
\label{sec:Dataset}
The Sen1Flood11~\cite{bonafilia2020sen1floods11} dataset is used for training and evaluation of the model. The dataset consists of 446 non-overlapped Sentinel-1 tiles. The samples are from 11 different flood events. Each sample is a patch of 512x512 pixels with 10-meter ground resolution. A wide variety of geographical areas are covered in the data, making it a good dataset for investigating the model's generalization capability.
Each sample is composed of two bands VV(vertical transmit, vertical receive) and VH(vertical transmit, horizontal receive). The dataset is also associated with Pixel-wise classification ground truth. Each pixel is classified into three categories, 0, 1, and -1. Class 0 represents the absence of water, class 1 represents water, and -1 indicates missing data.

The flood event samples in the Sen1Flood11 dataset are post-flood images. We strengthen the dataset by adding pre-flood images considering the Sentinel-1 images acquired with the same SAR geometry. We fetched the geometry and the orbit of the post-flood images, and downloaded all the available Sentinel-1 images over a span of 1 year before the flood event date. These Sentinel-1 images are downloaded using Google Earth Engine's python API ~\cite{GORELICK201718}. The pixel-wise median of all the past year images is considered as the pre-flood image.

The dataset is divided into training and validation sets as specified in the Sen1Flood11 dataset. The VV and VH backscatters of both pre and post-flood images are clipped in range (-23, 0)dB and (-28, -5)dB respectively. At last, all the images are normalized before feeding to the network. Few samples of pre-flood, post-flood images, and the corresponding flood mask are visualized in Figure\ref{fig1}.

\begin{figure*}  
\centering  
\begin{subfigure}
  \centering  
  \includegraphics[trim=1cm 0.2cm 1cm 1cm, width=165mm, height=70mm]{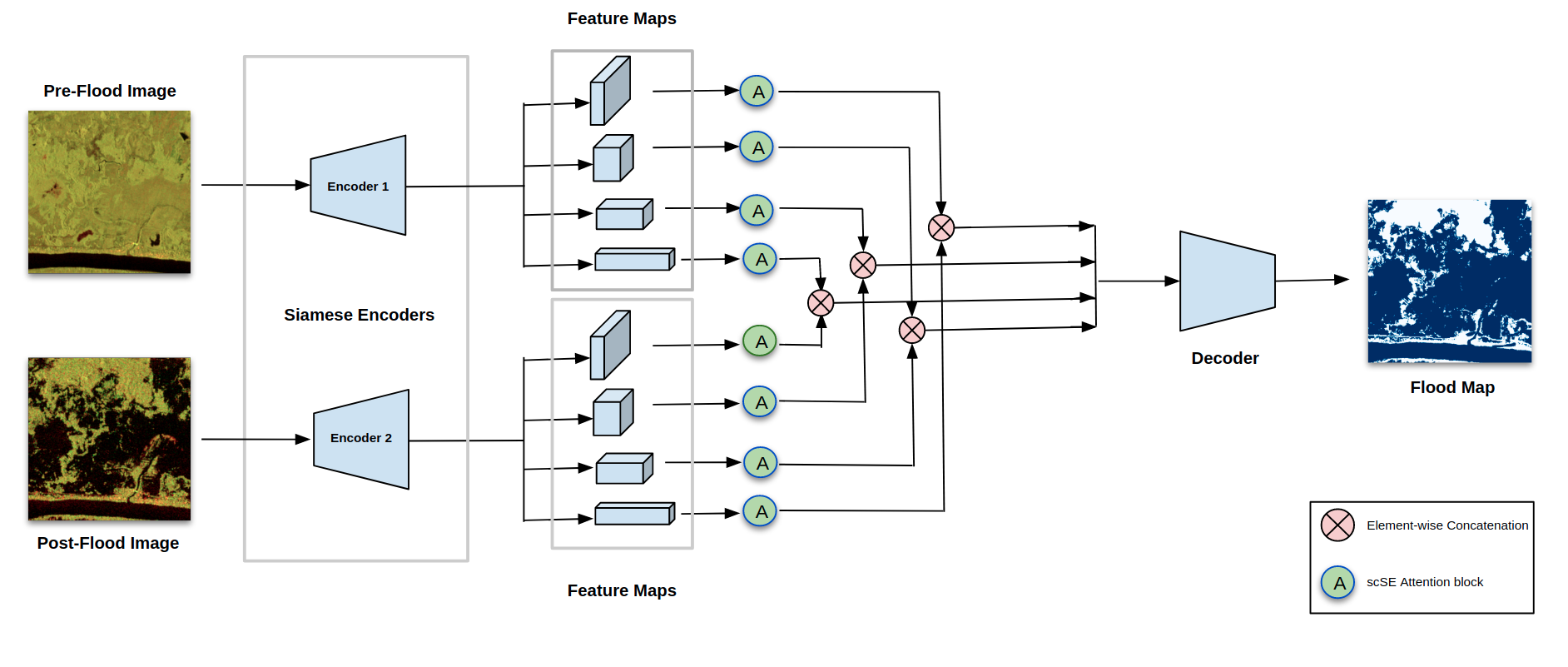}
\end{subfigure}
\caption{Attentive Dual Stream Siamese Network.}
\label{fig2}
\end{figure*}

\subsection{Network}
\label{sec:Network}
In this work, we propose a dual-stream Siamese network for flood detection. The network is shown in Figure\ref{fig2}. The architecture of the proposed network is inspired by the encoder-decoder segmentation networks. In such architectures, the encoder encodes the salient features of the input into a smaller representation named feature maps. These maps are then upsampled and decoded into a segmentation map in multiple steps. The size(width x height) of the segmentation map is equal to the size of the network's input. 

In the presented network, two encoders are used to encode pre-flood and post-flood images. The encoders used are inspired by siamese networks hence, share weights. The network takes 3 channel input, the first 2 channels are VV and VH SAR backscatter. The third channel is kept blank(all zeros). Since, We are using pre-trained 3 channel backbone, We are bound to use 3 channel input.

At different levels of the encoder, there are multiple-scale feature maps. At each scale, different level of semantic information is captured. From the two encoders, four feature maps of different scales are extracted. The size of the extracted feature maps are (256x256), (128x128), (64x64), and (32x32).

Since the dataset is acquired in regions with different terrain morphologies and land covers, the VV and VH backscatter behavior is not uniform. Depending on the surface, the significance of VV and VH channel varies. This phenomenon is taken into account by adding a channel-wise attention block to the network. The attention block is also complemented with the spatial attention and to achieve this we used Concurrent Spatial and Channel ‘Squeeze \& Excitation’(scSE) blocks~\cite{roy2018concurrent}.

The feature maps from the two encoders are now enhanced and weighted channel-wise. These features from the pre-flood and post-flood images are fused using concatenation operation. The feature maps are then fed into the decoder, where the output flood map is generated after applying, a series of convolution, upsampling, padding and normalization operations.

\subsection{Implementation and Training}
The pixel-wise change detection is handled as a binary classification task with two classes "change" and "no change". These two classes can be interpreted as flood and no flood. The problem of severe imbalance between changed and unchanged pixels is well known in the remote sensing field. To overcome this problem we used a combination of focal loss and dice loss to train the network. The loss combination used in the network is shown in equation(1).
\begin{equation} \label{Lapscore}
         Loss = \alpha * Dice Loss + (1-\alpha) * Focal Loss
         \end{equation}
For better convergence of the model, the learning rate is decayed in steps. The initial learning rate is 0.001 and decayed until it is at 0.00001. The decay steps are controlled with the "reduce on plateau" method, which decays the rate when the learning curve is stuck at a plateau. We conducted the experiments with multiple backbones and the best results are recorded with Resnet50 encoder.
In the learning process, 'GeoTIFF' images of size 512x512 pixels are used. With the help of augmentation, the data size and add geometric robustness to the model is increased. The augmentation methods used are horizontal and vertical flip. All the experiments are implemented in Keras and the network is trained on one google colab GPU. The network's training time is 2hours and inference time is 5 images per second. The Code will be made publicly available. 

\begin{figure*}  
\centering  
\begin{subfigure}
  \centering  
  \includegraphics[trim=1cm 1cm 1cm 1cm, width=150mm, height=100mm]{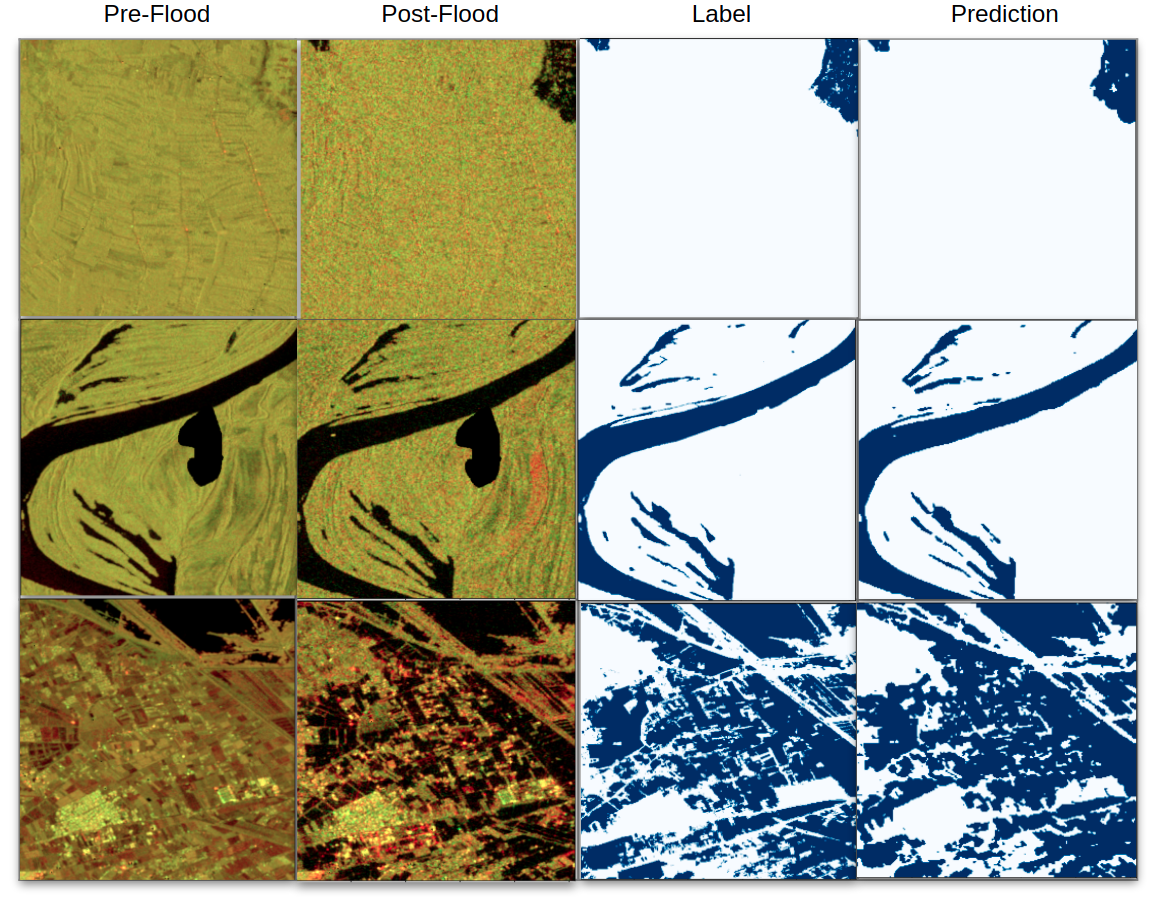}
\end{subfigure}
\caption{Detection Result Samples. Three sample results are visualized in 3 rows. From left to right: pre-flood, post flood images, Ground truth labels and proposed network's prediction are visualized.}
\label{fig3}
\end{figure*}

\section{Results and Evaluation}
\label{sec:eval}
For the quantitative evaluation of the proposed method, we used intersection over union(IOU) and F1-Score metrics. To the best of our knowledge, there are no works on this Sen1Flood11 or any available large-scale SAR dataset, which explore deep learning methods on multi-temporal data for flood detection.
There are two existing works conducted using uni-temporal post-flood data from the Sen1Flood11 dataset. Our results are compared with these methods referred here as 'DL Method 1' and 'DL Method 2'. Both of these methods work on post-flood data and the detection is performed as a segmentation task. The quantitative performance comparison is shown in Table \ref{SOTA_comp2}. 

\begin{table}[htbp]
\begin{center}
\resizebox{.9\columnwidth}{!}{%
\begin{tabular}{l|c|c}

\textbf{Methods} &\textbf{IOU} &\textbf{F1-Score} \\ \hline \hline

{Uni-Temporal DL Method 1\cite{bai2021enhancement}} &0.64 & --    \\ 
{Uni-Temporal DL Method 2 S1\cite{konapala2021exploring}}  & -- & 0.62   \\ 
{Bi-temporal Flood Detection\textbf{(ours)}}  & 0.70 & 0.83   \\  
\end{tabular}%
}
\end{center}
\caption{Performance comparison with existing methods.}
\label{SOTA_comp2}
\end{table}

From the comparison, we can see that our proposed method on multi-temporal SAR data outperformed the previous benchmark methods. The proposed method achieved 6\% better IOU in comparison to the 'DL Method 1' and 21\% better F1-score compared with 'DL Method 2'. A few samples of the flood detection results are visualized in Figure \ref{fig3} for qualitative analysis. The results prove that when the detection is done as a comparison between pre-flood and post-flood acquisitions, additional information is learned by the neural network improving the overall accuracy. This additional information contributes towards more accurate flood area detection.

\section{Conclusion}
\label{sec:conclusion}
In this work, we propose a dual-stream model to utilize pre-flood images along with post-flood images to detect the flood areas as a change detection task. From the evaluations, we found that with the help of pre-flood images, flood areas can be detected more accurately. Also, the Sentinel-1 data is freely available to download, hence utilizing before-event data adds no cost to the task and improves the flood detection results.In the next step, we will be extending our work to semi-supervised and unsupervised multi-temporal methods as labeled data are often not readily available and time-consuming to generate. We will aim to understand better the pros and cons in comparison to supervised methods. The over-reaching goal of our ongoing research is to provide robust and automatic methods for flood emergency mapping.

\bibliographystyle{IEEEbib}
\bibliography{refer}

\end{document}